\documentclass[a4paper, 11pt]{easychair}
\usepackage{graphicx}
\usepackage{amsmath}
\usepackage{amsfonts}
\begin{document}

%% Front Matter
%%
% Regular title as in the article class.
%
\title{On higher order computations, rewiring the connectome,  and non-von Neumann computer architecture }

% \titlerunning{} has to be set to either the main title or its shorter
% version for the running heads. When processed by
% EasyChair, this command is mandatory: a document without \titlerunning
% will be rejected by EasyChair

\titlerunning{On higher order computations, rewiring the connectome,  and non-von Neumann computer architecture}

% Authors are joined by \and. Their affiliations are given by \inst, which indexes
% into the list defined using \institute
%
\author{ Stanis\l aw Ambroszkiewicz 
%\and
 %  Graham Gough\inst{3}\thanks{Changed author list format.}\\
}

% Institutes for affiliations are also joined by \and,
\institute{Siedlce University of Natural Sciences and Humanities, Poland \and
Institute of Computer Science, Polish Academy of Sciences \\ 
%al. Jana Kazimierza 5, PL-01-248 Warsaw,    \\
  \email{sambrosz@gmail.com}
 }
%  \authorrunning{} has to be set for the shorter version of the authors' names;
% otherwise a warning will be rendered in the running heads. When processed by
% EasyChair, this command is mandatory: a document without \authorrunning
% will be rejected by EasyChair

\authorrunning{S. Ambroszkiewicz}

\clearpage

%%%%%%%%%%%%%%%%%%%%%%%%%%%%%%%%%%%%%%%%%%%%%%%%%%%
\maketitle
%%%%%%%%%%%%%%%%%%%%%%%%%%%%%%%%%%%%%%%%%%%%%%%%%%%
\pagestyle{headings}
\tableofcontents
\sloppy

%\setcounter{tocdepth}{2}
%{\small
%\tableofcontents}

%\section{To mention}
%
%Processing in EasyChair - number of pages.
%
%Examples of how EasyChair processes papers. Caveats (replacement of EC
%class, errors).

\pagestyle{empty}
%\date{April 14, 2015}

\begin{abstract}
Structural plasticity in the brain (i.e. rewiring the connectome) may be viewed as mechanisms for dynamic reconfiguration of neural circuits. First order computations in the brain are done by static neural circuits, whereas higher order computations are done by dynamic reconfigurations of the links (synapses) between the neural circuits.  
\\
Static neural circuits correspond to first order computable  functions. Synapse creation (activation) between them correspond to the mathematical notion of function composition. Functionals are higher order functions that take functions as their arguments. The construction of functionals is based on dynamic reconfigurations of  function compositions.  Perhaps the functionals correspond to rewiring mechanisms of the connectome. 
\\
The architecture of human mind is different than the von Neumann computer architecture. Higher order computations in the human brain (based on functionals) may suggest a non-von Neumann computer architecture, a challenge posed by John Backus in 1977  \cite{Backus}. 
The presented work is a substantial extension and revision of \cite{AmbroszkiewiczICANN2016}. 
 \end{abstract}

%\begin{keywords}
%  computations in human brain, higher order functions and functionals,  synaptic meta-plasticity, glia and atrocytes,  non-von Neumann computer architecture, Backus' function-level programming language
%Since the architecture of human brain is definitely not similar to  von Neuman computer architecture 1946, this may give rise to consider the rules that govern the meta-plasticity as a function-level programming language according to John Backus  1977
%\end{keywords}

\section{Introduction} 

{\bf Gedankenexperiment: a backward time travel of a computer.} 
%Time travel to the past:   
{\em A contemporary computer was  moved  into the XIX-th century so that scientists could make experimental research. Actually, the idea underlining the functioning of a computer is extremely simple; it is the von Neumann computer architecture. Would it be possible for the scientists of nineteenth century to discover the idea by examining the electric circuits and their  complex functioning of the working computer system consisting of monitor, a motherboard, a CPU, a RAM,  graphic cards, expansion cards, a power supply, an optical disc drive, a hard disk drive, a keyboard and a mouse? What about BIOS and operating system as well as many applications installed?} 

Perhaps the Gedankenexperiment may serve as a metaphor of the research on the human brain functioning.
 Although great discoveries and achievements have been made in Neurobiology, the basic mechanisms (idea) underling the human brain functioning are still a great mystery.

%%%%%%%  
A short review of the current research on  higher order computations in the brain is presented below.   
%WAŻNE raczej krótkie streszczenie pokazujące rolę glia w higher order computations in the brain 
Astrocytes are a kind of glial cells (glia). 
Let us cite the recent views on the role the glia play in meta-plasticity of the brain. 

Fields et al. 2015 \cite{Fields}: 
{\em ``Astrocytes have anatomical and physiological properties that can impose a higher order organization on information processing and integration in the neuronal connectome. 
Neurons compute via membrane voltage, but how do astrocytes compute? What do glia contribute to information processing that neurons cannot accomplish?
...  In comparison to neurons, glia communicate slowly and over broader spatial scales. This may make glia particularly well suited for involvement in integration, in homeostatic regulation, and alterations in structural or functional connectivity of neural networks taking place over periods of weeks or months.''}
%
% Yger and Gilson 2015  \cite{Ygier}: {\em ... metaplasticity is an ubiquitous mechanism acting on top  of classical Hebbian learning and promoting the  stability of neural function over multiple timescales, we stress the need for incorporating it as a key element in the framework of plasticity models. Bridging theoretical and experimental results suggests a more functional role for metaplasticity mechanisms than simply stabilizing neural activity.}
%

 Min et al. 2015 \cite{Min}: 
 {\em  ``Many studies have shown that astrocytes can dynamically modulate neuronal excitability and synaptic plasticity, and might participate in higher brain functions like learning and memory.   ... mathematical modeling will prove crucial for testing predictions on the possible  functions of astrocytes in neuronal networks, and to generate novel ideas as to how astrocytes can  contribute to the complexity of the brain. ...''}
 
Gilson et al. 2015 \cite{gilson2015editorial}: 
{\em ``Experiments have revealed a plethora of synaptic and cellular plasticity mechanisms acting simultaneously in neural circuits. How such diverse forms of plasticity collectively give rise to neural computation remains poorly understood.
...  To learn how neuronal circuits self-organize and how computation emerges in the brain it is therefore vital to focus on interacting forms of plasticity.''} 

Park and Friston 2013 \cite{park2013structural}:  
{\em `` ... the emergence of dynamic functional connectivity, from static structural connections, calls for formal (computational) approaches to neuronal information processing ...''}

 According to Bertolero,  Yeo, and D’Esposito (2015) \cite{bertolero2015modular}, so called  
{\em ``connector hubs''} are responsible for composition of modules (neuronal circuits) implementing cognitive functions.

Braun et al. (2015) \cite{braun2015dynamic}: 
{\em 
``... dynamic network reconfiguration forms a fundamental neurophysiological mechanism for executive function.''}   

The research on computational models of neural circuits is well established starting with  
McCulloch-Pitts networks \cite{McCulloch} via 
the Hopfield model (\cite{Hopfield1982} and \cite{Hopfield1986}) to recurrent neural networks (RNNs). It seems that RNNs  adequately represent  the computations done in the human brain by the real neuron networks. From the Computer Science point of view, RNNs are Turing complete (Siegelmann and Sontag \cite{Siegelmann}), i.e.,  every computable function may be represented as a RNN. 
However, Turing machine is a flat model of computation. There are also higher order computations, i.e. computable functionals  where arguments (input) as well as values (output) are functions.  For a comprehensive review of higher order computations, see Longley and Norman 2015 \cite{Longley}.  

The Virtual Brain (TVB \cite{SanzLeon},  www.thevirtualbrain.org) project aims at building a large-scale simulation model of the human brain. 
It is supposed that brain function may emerge from the interaction of large numbers of neurons, so that, the research  on TVB may  contribute essentially  to our understanding of the spatiotemporal dynamics of the brain's electrical activity. 
However, it is unclear how this activity may contribute to the comprehension of the principles of the human mind functioning.  
% jaki to ma sens? czy to cokolwiek wnosi do zrozumienia istoty działania mózgu? Czy to tylko bicie tzw. naukowej piany? 

Adolphs 2015 \cite{adolphs2015unsolved}: 
{\em ``Some argue that we can only understand the brain once we know how it could be built. Both evolution and development describe temporally sequenced processes whose final expression looks very complex indeed, but the underlying generative rules may be relatively simple ... ''}  

Another interesting approach is due to 
Juergen Schmidhuber: 
{\em ``The human brain is a recurrent neural network (RNN): a network of neurons with feedback connections''}; see   http://people.idsia.ch/~juergen/rnn.html . Indeed,  real  neural circuits can be modeled as (continuous time) RNNs.  
%Since synaptic metaplasticy was discovered, it is clear that  higher order RNNs are necessary. 
Despite the enormous complexity of a hypothetical RNN modeling the human brain, there is a paradox here because (continuous time) RNNs are nonlinear dynamic systems. It means that RNNs are high level mathematical abstractions (of human mind)  involving the notion of space-time Continuum that comprises actual infinity. These very abstractions are created  in the human brain (consisting of a finite  number of cells), i.e. the notions related to space-time continuum  are represented (in the brain)  as finite structures. 

 Some parts of the connecome may and should be considered as modules responsible for particular (elementary) cognitive functions of the brain. This very modularity  reduces considerable the complexity.  Once the modules are distinguished as functions with clearly defined input and output, it gives rise to compose them. The composition is, in turn, the basic mechanism for constructing higher order functionals. However, it seems that RNNs still lack the modularity and ability to compose the modules. Perhaps, if the notions of modularity and  computable functionals were introduced to RNNs, they could model the higher order computations as dynamic formation and reconfigurations of the links (synapses) between the neurons. 

Let us shortly review (in the form of citations) the current literature on the modularity in the human brain. 

 Bertolero et al. 2015 \cite{bertolero2015modular}: 
{\em ``The principle of modularity, in which a system or process is mostly decomposable into distinct units or “modules,” explains
the architecture of many complex systems. Biological systems, including the human brain, are particularly well explained by the
principle of modularity.''}

Sporns et al. 2016 \cite{sporns2016modular}: 
{\em ``Behavior and cognition are associated with neuronal activity in distributed networks of neuronal
populations and brain regions. These brain networks are linked by anatomical connections and engage in complex patterns of neuronal communication and signaling.''
}

Gu et al. 2015 \cite{gu2015controllability}: {\em ``Cognitive function is driven by dynamic interactions between large-scale neural circuits or networks, enabling behaviour. However, fundamental principles constraining these dynamic network processes have remained elusive.''}

Braun et al. 2015 \cite{braun2015dynamic}: {\em  ``The brain is an inherently dynamic system, and executive cognition requires dynamically reconfiguring, highly evolving networks of brain regions that interact in complex and transient communication patterns. However, a precise characterization of these reconfiguration processes during cognitive function in humans remains elusive.''}

From Editorial of van Ooyen and Butz-Ostendorf 2017 \cite{Ooyen}: 
{\em ``Exploring structural plasticity can be greatly assisted by mathematical and computational modeling. 
... 
Many findings have revealed that the adult brain is capable of altering its wiring diagrams, but the driving forces underlying structural plasticity and its role in brain function remain unclear.''}

Let us summarize the review. 

Recent advances in Neurobiology prove that reconfigurations of functional brain networks are responsible for higher  organization of information processing in the neuronal connectome; see also the following references that support this view; La Rosa et al. (2020) \cite{La_Rosa}, Bennett et al. (2018) \cite{Bennett}, Hilgetag et al. (2019) \cite{Hilgetag}, Nolte et al. (2020) \cite{Nolte}, Hearne et al. (2017) \cite{Hearne}, Bassett and Sporns (2017) \cite{Bassett}, Marek et al. (2019) \cite{Marek}, Dixon et al. (2018) \cite{Dixon}, and Leopold et al. (2019) \cite{Leopold}.   

A conclusion from the above papers is that generic mechanisms for dynamic reconfigurations of connections between neural circuits are not known. 
Since information processing in the brain is identified with computations performed by neural circuits, higher organization of information processing in the brain corresponds to higher order computations in Computer Science.  
They are based on the notion of functionals that take functions as arguments and return functions as values. 
It is well known that constructions of such functionals are based on dynamic creations and reconfigurations of data-flow connections between elementary functions. 
This fact strongly relates functionals to the generic mechanisms for dynamic reconfigurations of connections between neural circuits in the brain. 

The foundations of the mind functioning might be ingenious in its simplicity although the underlying biological mechanisms are extremely complex and sophisticated.   
Hence, in order to model neural circuits and the mechanisms responsible for structural changes in the neuronal connectome, let us use much more simple (than RNN) primitive notions from Mathematics and Computer Science, i.e. the computable functions and computable functionals.  Since Mathematics is a creation of the human mind,  the Foundations of Mathematics may shed some light on the principles of the brain functioning. That is, the basic mathematical notions can be recognized as concrete mental structures, and then the corresponding  mechanisms of the human brain can be discovered. 

Evidently, the architecture of human mind is different than von Neumann computer archtecture. 
Perhaps a non-von Neuman architecture (postulated by John Bacus \cite{Backus}) may result in implementation of the functionals in hardware in the very similar way as it is done in the human brain. We will explore this idea in Section \ref{CGRA}.  

%% 
%%%%%%
\section{Neural circuits, computable functions and functionals} 

\begin{figure}[h]
	\centering
	\includegraphics[width=0.7\textwidth]{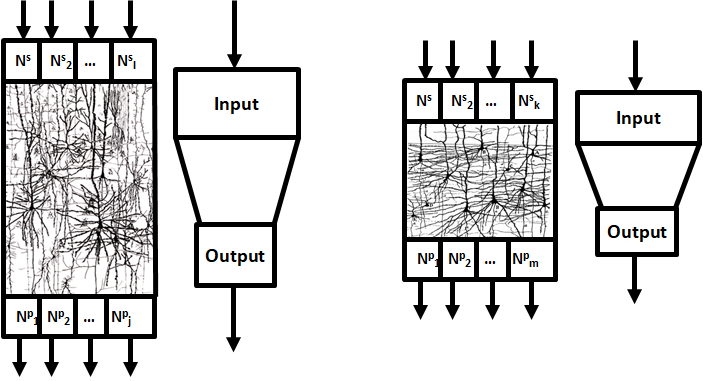}
	\caption{Abstraction: neural circuits as functions. A function consists of input (socket), body, and output (plug) }
	\label{elementary_cognitive_functions_as_neuronal_circuits}
\end{figure} 
Before going into details, several assumptions are to be made. The first one is that elementary neural circuits (corresponding to functional units of the brain) can be distinguished. The second assumption is that any such circuits (at least temporary)  has clearly identified input (dendrite spines of some postsynaptic neurons) and output (axons of some presynaptic neurons). It means that the output is exactly determined by the input. 
The third assumption is that such circuits can be composed by a linking  the output of one circuit to the input of another circuit; it may be done by  (activating) creating a (silent) synapse connecting an axon (of the output of one circuits) to a dendrite spine of the input of the other circuit. If the above assumptions can be verified experimentally, then the following considerations make sense. However, from the conceptual point of view, they may also be of some interest to Neurobiology.  

If the above assumption are taken as granted, then a neural circuit can be represented as a first order function defined on natural numbers, see Fig. \ref{elementary_cognitive_functions_as_neuronal_circuits}. That is, spike sequences (bursts), generated by a neuron,  may be interpreted as natural numbers in the unary code, input of the circuit as arguments whereas output as values of the function.  Note, that this is a static (one shot) representation of neural circuits. It means that one output is produced form one input. 

However, if a circuit is to be considered in a time extent so that for consecutive inputs, it produces a sequence of outputs, then dynamic behavior of the circuit may be represented either as a RNN or as a sequence of interrelated copies of the function representing the circuit; for details see \cite{TO}. 

\begin{figure}[h]
	\centering
	\includegraphics[width=0.7\textwidth]{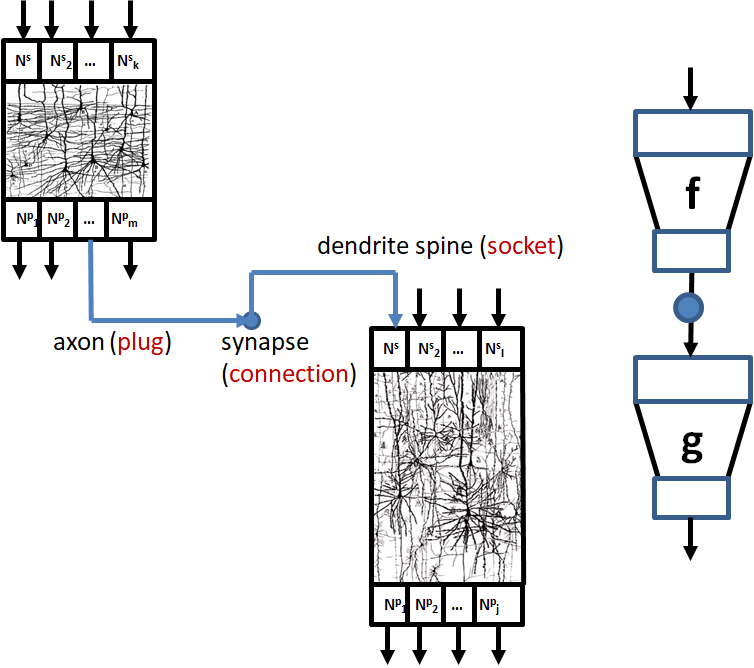}
	\caption{Connection as composition of two functions }
	\label{Connection_as_function_composition-one_of_the_basic_notions_in_Mathematics}
\end{figure}

Simple operations on functions may have their counterparts as operations on neural circuits. Given two functions $f$ and $g$ (from natural numbers into natural numbers), the new function $h$ defined as $h(x) = g(f(x))$ may serve as  an example, see Fig. \ref{Connection_as_function_composition-one_of_the_basic_notions_in_Mathematics}. If $f^c$ and $g^c$   denote corresponding neural circuits, then the circuit corresponding to function $h$ may be created by establishing (activating) synapses between input neurons of $g^c$, and the output neurons of $f^c$. This may correspond roughly to the synaptic meta-plasticity. It is interesting (however, not surprising) that this very synapse creation corresponds to a basic notion of Mathematics, i.e. to the function composition. 

{\bf Sockets} and {\bf plugs} are the crucial notions. A function consists of input, body and output, see Fig. \ref{elementary_cognitive_functions_as_neuronal_circuits}.  Input may consists of multiple sockets, whereas output may consists of multiple plugs. 
A plug-socket directed link may correspond to synapse as connection of axon and dendrite.   

There are also higher order functions (called functionals) where arguments as well as values may be functions. It is also not surprising that these higher level functionals can be constructed by establishing links between plugs and sockets. 

Each function is of some type. Since the natural numbers (finite sequences (bursts) of spikes) are assumed as the basic type (denoted by $N$), a type of first order functions is of the form 
$$
(N^{s_1}; \ N^{s_2}; ... ;N^{s_k})  \rightarrow (N^{p_1};\ N^{p_2}; ... ;N^{p_m})
$$ 
where $(N^{s_1}; \ N^{s_2}; ... ;N^{s_k})$
denotes different sockets of the input, whereas \\ $(N^{p_1};\ N^{p_2}; ... ;N^{p_m})$ denotes different plugs of the output.  This type may be realized as a board consisting of sockets and  plugs. %, see Fig. \ref{nc-12}. 

 %%%%%%%%
 %\subsection{Higher order computations } 
It seems that second (and higher) order computations in the brain are done by dynamic (re)configurations of  links (synapses)  between the neural circuits. Although the links are established between concrete neurons, these neurons belong to fixed circuits, so that (from functional point of view) the links are between circuits and correspond to circuit compositions. 
 
Let us take as granted that higher order computation in the human brain is based on dynamic creating, composing, and reconfiguring neural circuits.  At the bottom level it is realized by creating new synapses or activate existing silent synapses (see e.g. Kanold et al. 2019 \cite{Kanold}); this corresponds to function composition. Since the function composition is the basis for construction of the higher order functions (functionals), the mechanisms responsible for dynamic synapse creation and activation correspond (can be abstracted) to functionals. 

Actually, function composition can be abstracted to a second order function (functional) that takes (as input) two  first order functions, a plug of one function and a socket of the second function; then it returns (as the output) a first order function as a composition of these two functions.  
  
We may wonder how such generic composition is realized in the brain. First of all, the circuits to be composed must be discriminated,  and then passed, as parameters, to the composition mechanism. %It is interesting how the discrimination is realized in the brain. 

Functionals (by their construction) have hierarchical structure that might correspond to the hierarchical organization of cortical and thalamic connectivity, see  Harris et al, (2019)  \cite{harris2019hierarchical}. 
Hence, the following hypothesis seems to be justified. 

 {\bf  The primitive  rules for construction of the computable functionals may have their counterparts in the human brain. } 
 
%
%%%%%%%%%%%%%
 \subsection{A sketch of formal framework for constructing higher order computation based on functionals } 
 
Turing machines and partial recursive functions are not concrete constructions. Their definitions involve actual infinity, i.e. infinite type for Turing machines, and minimization operator $\mu$ for partial recursive functions. This results in possibility of {\em non terminating computations} that are abstract notions and have no grounding in the human brain. 

\begin{figure}[h]
	\centering
	\includegraphics[width=0.6\textwidth]{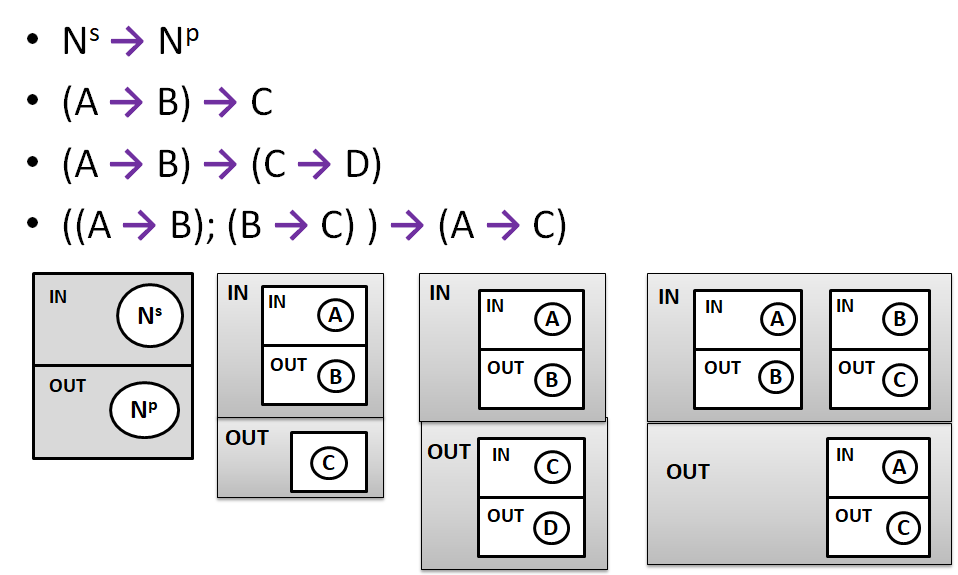}
	\caption{Function type as board of sockets and plug. Higher order types as nested boards of sockets and plugs }
	\label{Sockets_plugs_and_higher_order_types}
\end{figure} 
\begin{figure}[h]
	\centering
	\includegraphics[width=0.85\textwidth]{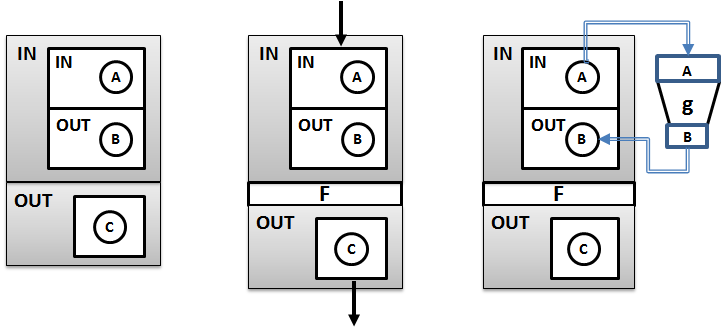}
	\caption{ Higher order application of functional $F$ of type $(A \rightarrow B) \rightarrow C$ to a function $g: A\rightarrow B$. The result $F(g)$ is an object of type $C$ }
	\label{nc-34}
\end{figure} 
The proposed approach is fully constructive, and if restricted only to first order computable functions, it correspond to the general recursive function according to the Herbrand-G\"{o}del definition. 

At the basic level it consists of some primitive types, primitive functions and type constructors, i.e. the type of natural numbers, the  successor function, constant functions, projections, constructors for product and function type. The key primitive functionals correspond to application, composition, copy and iteration.  It is crucial that these functionals can be constructed by (dynamic, in the case of iteration) establishing links between plugs (corresponding to output types) and sockets (corresponding to input types). 

At the higher level of the approach, types are considered as objects, i.e. constructed as {\bf boards of plugs and sockets}. This gives rise to introduce relations (according to the propositions-as-types correspondence of Curry-Howard), and polymorphism.  

Hence, it is important to grasp the constructions of the boards as higher order types.  The type of functions from natural numbers into natural numbers (denoted by $N^s \rightarrow N^p$) may be realized as a simple board consisting of a socket and a plug, see Fig. \ref{Sockets_plugs_and_higher_order_types} where also types of higher order are presented.  Note that for the type $(A \rightarrow B) \rightarrow C$,  the input $A \rightarrow B$ becomes the socket. For the type $(A \rightarrow B) \rightarrow (C \rightarrow D)$,  the output $C \rightarrow D$ becomes the plug.

 \begin{figure}[h]
	\centering
	\includegraphics[width=0.99\textwidth]{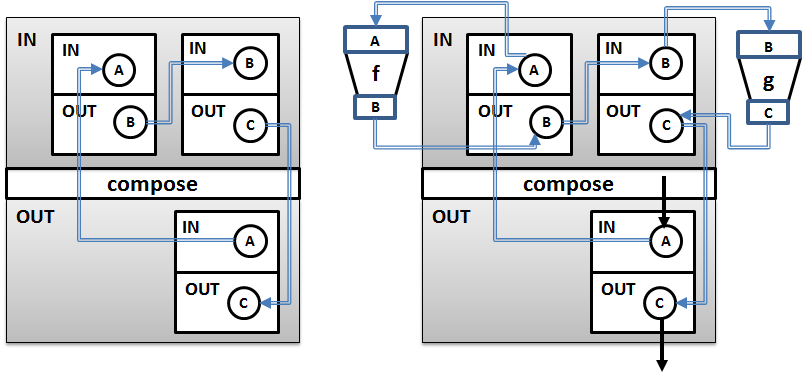}
	\caption{ The functional $Comp$ of type $((A \rightarrow B); (B \rightarrow C)) \rightarrow (A \rightarrow C)$. Input objects are: $f$ of type $A\rightarrow B$, and $g$ of type $B\rightarrow C$.  When applied to $Comp$, the output object is a function of type  $A\rightarrow C$  }
	\label{nc-5}
\end{figure} 

{\bf Application} of a functional $F: (A \rightarrow B) \rightarrow C$ to a function $g: A \rightarrow B$ is realized as follows.  $A\rightarrow B$ is the socket of the functional $F$. The application is done (see Fig. \ref{nc-34}) by establishing appropriate directed connections (links). That is, the link between the socket  $A$ of the socket of $F$ and the socket $A$ of $g$, and the link between the plug $B$ of $g$ and the plug of the socket of $F$.

{\bf Composition functional} (denoted by $compose_{A,B,C}$) for simple composition of two functions (the first function $f$ of type $A\rightarrow B$, and the second one $g$ of type $B\rightarrow C$) is realized as two boards with appropriate links shown in Fig. \ref{nc-5}. It is easy to check (by following the links) that applying $compose_{A,B,C}$ to two functions (see Fig. \ref{nc-5}) results in their composition. 

Note that a higher order application (i.e. application of a functional to a function), and  a functional for composition are constructed just by providing some links between sockets and plugs. Since link corresponds to synapse, it might be interesting whether these functionals have counterparts in the brain. 

Each construction, like $F(g)$ and $compose_{A,B,C}(f;g)$, can be distinguished as an individual object (notion). Perhaps, in the brain, they correspond to concrete regions. 
This corresponds to a new paradigm called radical embodied neuroscience (REN), see Matyja and Dolega 2015 \cite{Matyja}, Kiverstein and Miller 2015 \cite{Kiverstein}. 

Generally, discrimination of new notions by the human mind is crucial for reasoning.  Once a notion is distinguished, it may be used in more sophisticated reasoning. This evidently corresponds to the {\em reflective abstraction} introduced by Piaget, especially if the notions emerge as the results of constructions. Note that here {\em constructions} mean  dynamic (re)configuration of links between sockets and plugs. 

A functional of special interest is $Copy$.  Once an object $a$ is constructed, repeat the construction once again. So that  $Copy(a)$ returns two object: the original $a$, and  its copy $a'$. Although the meaning of $Copy$ seems to be simple, its realization in the brain may be quite complex especially if the object $a$ is of a higher order type.  

If it is supposed that the construction of object $a$ occupies some well defined region in the brain, then  $Copy$ may be realized by copying this region into a new ``free region''.  
Since in Biology (living organisms) copying (procreation) is ubiquitous, let us take the implementation of the functional $Copy$ as granted.  
\begin{figure}[h]
	\centering
	\includegraphics[width=0.99\textwidth]{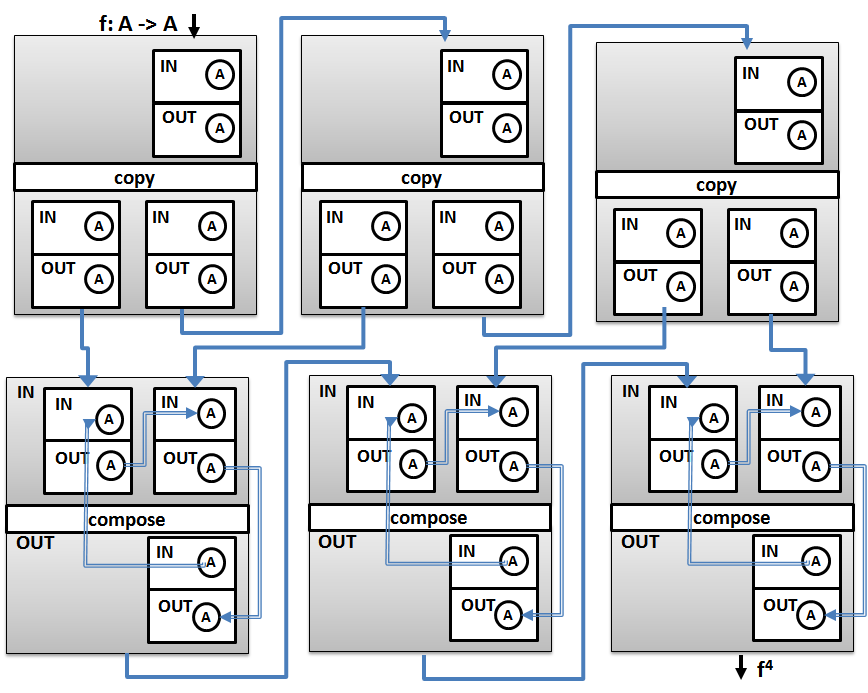}
	\caption{ The result of application of the functional $Iter_A$ to natural number $4$ and function  $f$ of type $A\rightarrow A$  }
	\label{nc-6}
\end{figure} 

{\bf Iteration as generalization of composition.} That is, compose $n$-times a function $f:A \rightarrow A$ with itself. Note that $n$ as a natural number is a parameter. The iteration is denoted by $Iter_A$ and it is a functional of type $(N; (A \rightarrow A)) \rightarrow (A \rightarrow A)$. So that  $Iter_A(n;f)$ is the function being $n$-time composition of $f$. The realization of   $Iter_A$ requires $Copy$ for making copies of $f$, and $(n-1)$ copies of the composition functional, see Fig. \ref{nc-6}, where the construction is done for $n$ equal $4$. Since natural numbers are involved in the functional, it seems that a hypothetical   realization of $Iter$, in the brain, requires neurons.

Functional $Iter$ is not the same as feedback loop that occurs when outputs of a circuit are routed back as inputs to the same circuit.  The feedback enforces dynamics of the circuits, whereas $Iter$ is static one shot operation. 

Feedback loop can not be realized for higher order functionals where input as well as output are not electrical signals but higher order constructions.   

Neural circuits are real dynamic systems where computation is done by consecutive processing signals (spike bursts). The circuits may be represented statically (without dynamics) as first order functions.  Functionals are also static constructions operating on first order functions (circuits) by re(configuring) links inside and between the circuits. 

Higher order primitive recursion schema (also known as Grzegorczyk's iterator) can be constructed as a functional. 
For arbitrary type $A$, the iterator, denoted by  $R^A$, of type $A\rightarrow ( (N\rightarrow (A\rightarrow A)) \rightarrow  (N\rightarrow A) )$, is defined by the following equations. 

For any $a:A$, \ \ $c:N\rightarrow (A\rightarrow A)$, and $k:N$

$((R^A(a))(c))(1) = a $ \ \ \  and \ \ $((R^A(a))(c))(k+1) = (c(k))(((R^A(a))(c))(k))$

However, a construction of $R^A$ does not follow from the definition. Actually, it is based on the iteration functional and consists on dynamic formation of links in boards of plugs and sockets. Higher order primitive  recursion allows to define a large subclass of general recursive functions, e.g. the famous Ackerman function.  This can be done on the basic level of the proposed approach to computable functionals. At higher levels of the approach (where functionals are used) all general recursive functions can be constructed. It seems that higher order computation involving the functionals is useful, especially as efficient and smart organizations of complex and sophisticated first order computations.

Note that there are next orders of constructions of functionals. Functionals operate on functionals (second order functions) are third order functions that operate on the second order functions by re(configuring) links in the boards of sockets and plugs.    By analogy, this may be continued for the next higher orders  constructions. 

The complete presentation of the approach for constructing functionals is done in \cite{TO}. It is a strictly mathematical point of view. 

Another point of view (close to the connectome) is considered in Section \ref{CGRA}, where functionals are related to generic mechanisms for dynamic  reconfigurations of compositions of first order functions. There, functionals are useful abstractions that reduce complexity of management of function compositions in large arrays of first order functions. Such array may be interpreted as a connectome structured by elementary neural circuits with   connections between them that can be activated or deactivated. 

%%
%%%%% dotąd 3 października 2020
\section{Continuum }

It seems that the notion of continuum has a straightforward and natural grounding in the human brain. 

{\bf Vision sensory nervous system.} 
The retina consists of about 130 million photo-receptor cells, however, there are roughly 1.2 million axons of ganglion cells that transmit information from the retina to the brain. It is interesting that a significant amount of visual pre-processing is done between neurons in the retina.  The axons form the optic nerve consisting of fibers (axons). 
Positions of the fibers in the nerve reflect the spatial and adjacent relations between the corresponding photo-receptors in the retina. In computations, the bundle of spikes in the nerve is considered together with the adjacent relation between the spikes.  It is crucial for comprehending the notion of space Continuum. 

{\bf The somatosensory system.} Contrary to the vision system, it  is spread through all major parts of mammal's body. Spacial and adjacent relations between nerve fibers of the somatosensory system contribute essentially to the notion of space Continuum. 

The streams of spikes, in the nerve fibers delivered from sensory receptors to the brain, are not independent form each other; they are structured by causal and adjacent relations. The streams along with the relations are the grounding for the notion of space-time Continuum. 

 It is interesting that objects of the primitive types are based on neuron spikes. Natural number is just an isolated independent spike burst, whereas object of the type Continuum is a bundle of adjacent spike bursts.  

A mathematical approach to the grounding (meaning) of the notion of Continuum is proposed in \cite{C}. In the context of Neurobiology,   of some interest may be also constructive foundations of Geometry in \cite{geometries}.

%%%%
%%%%%%%%%%%%
\section{CGRA and connectome } 
\label{CGRA}

% zmień bo to żywcem wzięte z TO
 A Coarse Grain Reconfigurable Architecture (CGRA) Bjorn De Sutter et al. (2013) \cite{de2013coarse}, is a type of processor architecture that can be reconfigured at runtime. Array of interconnected functional units (first order functions implemented in hardware) can be dynamically configured  (composed). 
Here, functionals may be envisioned as generic mechanisms for the management of connections (compositions) between the functions  in the reconfigurable arrays. 
Hence, the concept of higher order computations as dynamic configuration of connections between hardware functional units is worth to be explored. 

Reconfigurable computing architectures (Russell Tessier et al. (2015)  \cite{tessier2015reconfigurable}), and reconfigurable system  (James Lyke et al. (2015)  \cite{lyke2015introduction}) are active research subjects. However, the correspondence between functionals and generic mechanisms for dynamic reconfigurations is still not recognized in the hardware design.

 % Perhaps the most advanced approach was done by Dan Ghica et al. 2011 \cite{ghica2011geometry} that is limited, however, to compiling affine recursive programs.    

%Software and hardware are two different worlds. The first world consists of tightly coupled software (von Neumann programming languages) and extremely simple and ingenious von Neumann computing architecture. Computation is done on bytes sequentially on a single processor. 
%The second world is hardware with its rich diversity of possible computing architectures, and possible grounding for higher order computations on functionals. 

%A new concept of programming (different form the von Neumann languages) is needed to shift the contemporary computing paradigm. Higher order computation can be envisioned as dynamic creation and reconfiguration of links between elementary circuits representing first order functions. 

% Perhaps the answer is in technology for developing large dynamically reconfigurable arrays of integrated circuits representing first order functions.  

CGRA is an active research area. New concepts are explored like:  data-flow graphs ( Niedermeier et al. (2014) \cite{niedermeier2014dataflow}, and Palumbo et al. (2016) \cite{palumbo2016dataflow}), full pipelining and dynamic composition  (Cong et al. (2014) \cite{cong2014fully}), and overlay architecture  (Capalija et al. (2014) \cite{capalija2014tile}, Jain et al. (2016)  \cite{jain2016adapting}, and Andrews et al. (2016) \cite{ma2016just}). 

Functional units (FUs), as elementary first order functions, are built on fine grained integrated circuits. FUs are collected in an array, and connections between them are reconfigurable. A program (to be realized in hardware) is modeled as a data-flow graph where the nodes correspond to FUs. If the array is sufficient rich in FUs and possible connections, the graph can be mapped into the array. If the graphs are acyclic, then the data flow is fully pipelined, and can be realized in hardware, i.e. in the array of FUs.   
 
Fully pipelined data-flows (as directed acyclic graphs) correspond to simple programs. Sophisticated programs may use recursion that presupposes dynamic unfolding, i.e.  dynamic transformation of acyclic graphs, during execution, see Mukherjee et al. (2017) \cite{Mukherjee}. 

The transformation corresponds to a generic mechanisms for dynamic configuration of connections in large arrays of FUs.  
If the number of the connections is at most dozens (for simple computations), then dedicated mechanisms may be designed. However, if hundreds, thousands, and even more connections and FUs are needed, then the mechanisms must be generic, and must be based on higher order abstractions, i.e. higher order types and higher order functionals. 

Hence, a functional is a fully pipelined data-flow (directed acyclic graph) with nodes corresponding to primitive functionals, and edges corresponding to flow of data from output of one node to the input of the next node.
For complex functionals (with recursion) some nodes are dynamically unfolded. 

From abstract mathematical point of view, functional is dynamic (depending on input values) transformation of acyclic directed graphs. 

A functional is executable (is an executable program) if all input and output nodes of the graph are first order functions. This means that the functional (dynamically) configures compositions of the first order functions. 

Executive programs can be compiled into hardware i.e. into an underling array of FUs.

%%%%%%
%%%%%%%%%%%%
\section{A non-von Neumann architecture (of human mind?)}
\begin{figure}[h]
	\centering
	\includegraphics[width=0.4\textwidth]{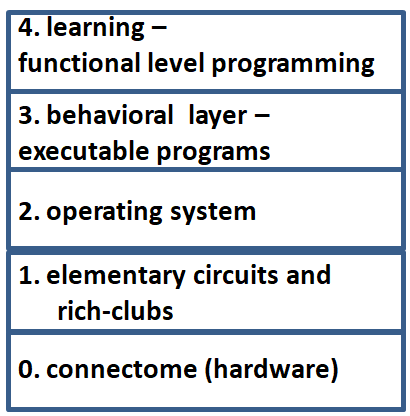}
	\caption{Layered architecture of a non von Neumann architecture (of human mind?)  }
	\label{layers}
\end{figure} 

Let us recall that functionals are constructed by dynamic creation and reconfiguration of links between sockets and plugs in the very similar way as the  higher order organization of information processing in the neuronal connectome. 

Links (connections) are between plugs and sockets of the same type. Although the type may be of higher order, they can be reduced to the links between sockets and plugs of the primitive type of natural numbers. 

A link is always directed, i.e. it determines the direction of data flow. 

We are going to propose a layered architecture of the human brain, see Fig. \ref{layers}. 

The basic layer is the connectome, i.e. neurons connected into circuits. It corresponds to hardware in the von Neumann computer architecture. It is 0-layer. 

% Each of the consecutive layers is an abstractions of the lower layer. 

Let us suppose, for a moment, that the connectome can be represented as a repository (array) of elementary circuits corresponding to first order functions. The array is equipped with rich-clubs, silent synapses (that can be activated), and potential new synapses that can be created on demand. 
This constitutes the 1-st layer of the architecture. 

\begin{figure}[h]
	\centering
	\includegraphics[width=0.99\textwidth]{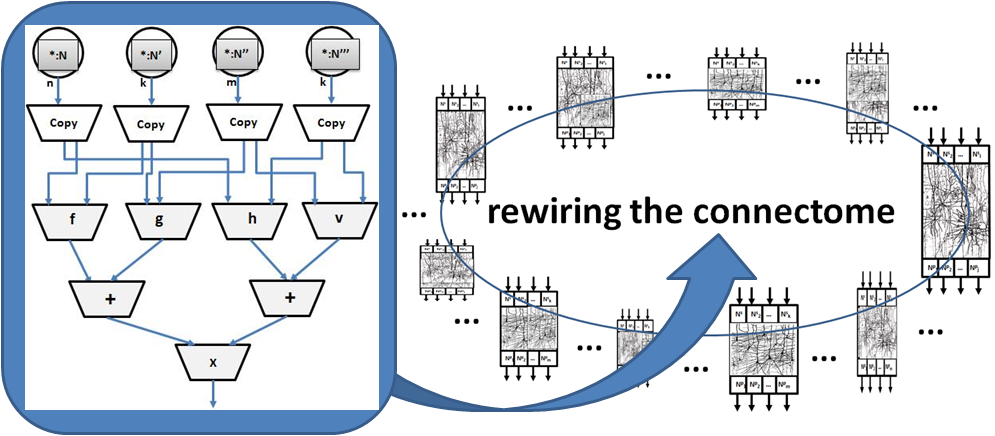}
	\caption{Executable functionals (programs) implemented in connectome  }
	\label{data_flow_to_connectome}
\end{figure}

\begin{figure}[h]
	\centering
	\includegraphics[width=0.7\textwidth]{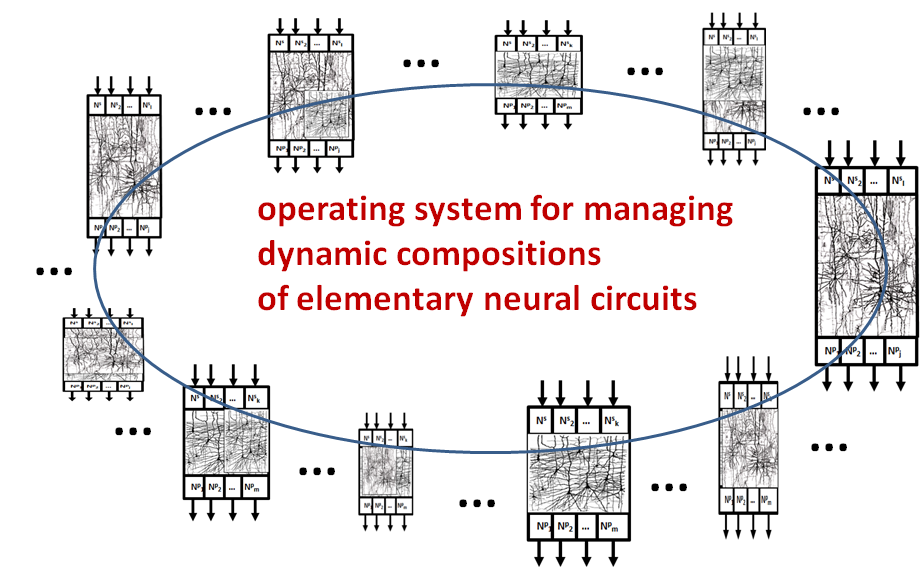}
	\caption{Operating system for managing  dynamic reconfigurations of compositions of first order functions (i.e. elementary neural circuits) }
	\label{operation_system_for_management}
\end{figure}

% executable program: if all input and output nodes of the graph are first order functions. 
In order to compile an executive functional (program) into the repository (see Fig. \ref{data_flow_to_connectome}), an operating system is needed for managing dynamic compositions of elementary neural circuits into a complex circuit, see Fig. \ref{operation_system_for_management}. 
This constitutes the 2-nd layer of the architecture.
Executive programs constitute the 3-rd layer. It is called behavioral layer where, triggered by external (or internal) stimuli,  executable programs are passed to the operating system (layer 2) to be compiled in layer 1 onto complex dynamic neural circuits. 

Any such executive program is constructed from complex higher order functionals, the functionals that are constructed from  primitive functionals according to generic rules. It is done in the 4-th layer called functional level programming layer. Construction of new functionals is related to learning in a general sense. Hence,  the layer is called learning.  
This may be seen as higher order programming, i.e. constructions of higher order functionals.  Hierarchy of such functionals corresponds to hierarchy of abstract notions. 

Note that operating system, behavioral layer and learning are consecutive and hierarchical abstractions done above the physical layers of connectome with elementary circuits and rich-clubs.  
One may ask where and how these abstract layers are implemented in the brain. 

% the result of program execution is not a value as it is in von Neumann. It is a complex circuit.  

%Surprising is that the repository of elementary cognitive functions (neural circuits) is limited to a number that is not so large, comparing to the capacity (memory) and capability (RAM and CPU speed) of contemporary computers based on von Neumann architecture. 
% 

%%
%%%%%%
\section{Conclusion}  

Leopold et al. (2019) \cite{leopold2019functional}:
{\em ``One of the great challenges in the study of the brain is to synthesize a large number of details into principles for understanding. While many details are known about the cerebral cortex, our level of understanding about its overarching architectural and functional principles remains, arguably, primitive.'' }

Finally, let us recall the fundamental assumption that can relate functionals (as higher order computing) to higher order information processing in the human brain. 

% The two functionals ($Copy$ and $Iter$) together with the higher order application, composition, and the primitive types constitute the cornerstone for building a constructive (intuitionistic) part of Arithmetics and Analysis, see \cite{TO} and \cite{C}.  According to the original meaning  of  L. E. J. Brouwer, intuitionism is the constructive mental activity of the human mind. 

It seems that there are two essential primitive types in Mathematics; the type of natural numbers, and the type of Continuum. Both types have their counterparts in the human brain. The natural numbers may be identified with  individual and independent bursts of neuron spikes. The type Continuum has also the straightforward interpretation in the human brain. Sensory nervous systems support this view. The static interpretation of the neural circuits, as first order computable functions, seems to be justified. This may give rise to expect that higher order computable functions (functionals) have counterparts in the human brain. 

Composition (as link creation) is the basic operation for function constructions as well as for construction of higher order functions (functionals). This very composition corresponds to synapse creation (activation) in the brain. 

Functionals are construction (as mental structures) of the pure intellect of the human mind. 

Hence, the following hypothesis seem to be reasonable:  \\ {\em\bf  The mechanisms that are responsible for the higher order organization of information processing in the human brain correspond to computable functionals. }

There are experimental evidences confirming the hypothesis, e.g. Bertolero (2015) et al. \cite{bertolero2015modular}, Braun et al. (2015) \cite{braun2015dynamic}, and Park and Friston (2013)  \cite{park2013structural}.  

The principles for constructing such mechanisms (functionals) are extremely simple like the ones that are basis for the von Neumann computer architecture. Contemporary computers (built on the the von Neumann computer architecture) are sophisticated in their functionality. In the same spirit, human brain and its functioning (self-consciousness, cognitive structures, behavior, learning, emotions, feelings, relation with other humans, etc.) are sophisticated despite the fact that they are based on extremely simple principles. 

Since the architecture of human brain is definitely different than  von Neumann computer architecture (see  von Neumann 1958 \cite{vonNeumann} and 1966 \cite{vonNeumann1966automata}),  the mechanisms for rewiring the connectome  (i.e. the meta-plasticity) may give rise to develop a non-von Neumann computer architecture and a corresponding function-level programming language postulated by John Backus  1977 \cite{Backus}; for more on this subject see \cite{HOF}. 
 
\nocite{*}
 \bibliography{NC}
\bibliographystyle{fundam}
%\bibliography{bibliofi} %% BibTeX

\end{document}